\documentclass{article} 
\usepackage{conf,times}

\usepackage{amsmath,amsfonts,bm}

\def\eqref#1{equation~\ref{#1}}
\def\Eqref#1{Equation~\ref{#1}}

\def\1{\bm{1}}

\DeclareMathAlphabet{\mathsfit}{\encodingdefault}{\sfdefault}{m}{sl}
\SetMathAlphabet{\mathsfit}{bold}{\encodingdefault}{\sfdefault}{bx}{n}

\usepackage{url}
\usepackage{graphicx}
\usepackage{xspace}
\usepackage{enumitem}
\usepackage{wrapfig}
\usepackage{booktabs} 
\usepackage{multirow}
\usepackage{algorithm}
\usepackage{algorithmic}
\usepackage{lineno}

\usepackage{color}
\usepackage{xcolor}
\definecolor{citecolor}{HTML}{3498DB}
\definecolor{linkcolor}{HTML}{E74C3C}
\usepackage[pagebackref=false,breaklinks=true,colorlinks,bookmarks=false,citecolor=citecolor,linkcolor=linkcolor]{hyperref}

\title{MetaMorph: Learning Universal Controllers with Transformers}

\author{Agrim Gupta$^1$,\: Linxi Fan$^{1,3}$,\: Surya Ganguli$^{1,2}$,\: Li Fei-Fei$^{1,2}$\\
$^1$Stanford University, $^2$Stanford Institute for Human-Centered Artificial Intelligence \\
$^3$NVIDIA Corporation \\

\texttt{\{agrim,sganguli,feifeili\}@stanford.edu}, \:\texttt{linxif@nvidia.com}\\
}

\newcommand{\maple}{MetaMorph\xspace}
\newcommand{\figvspace}{-4mm}

\newcommand{\mbf}[1]{\mathbf{#1}}
\newcommand{\op}[1]{\operatorname{#1}}
\newcommand{\codecomment}[1]{{\color{gray}{#1}}}
\iclrfinalcopy 
\begin{document}

\maketitle

\begin{abstract}
Multiple domains like vision, natural language, and audio are witnessing tremendous progress by leveraging Transformers for large scale pre-training followed by task specific fine tuning. In contrast, in robotics we primarily train a single robot for a single task. However, modular robot systems now allow for the flexible combination of general-purpose building blocks into task optimized morphologies. However, given the exponentially large number of possible robot morphologies, training a controller for each new design is impractical. In this work, we propose MetaMorph, a Transformer based approach to learn a universal controller over a modular robot design space. MetaMorph is based on the insight that robot morphology is just another modality on which we can condition the output of a Transformer. Through extensive experiments we demonstrate that large scale pre-training on a variety of robot morphologies results in policies with combinatorial generalization capabilities, including zero shot generalization to unseen robot morphologies. We further demonstrate that our pre-trained policy can be used for sample-efficient transfer to completely new robot morphologies and tasks. 
\end{abstract}

\section{Introduction}
\label{section:intro}
The field of embodied intelligence posits that intelligent behaviours can be rapidly learned by agents whose morphologies are well adapted to their environment \citep{brooks1991new,pfeifer2001understanding,bongard2014morphology,gupta2021embodied}. 
Based on this insight, a robot designer is faced with a predicament: should the robot design be task specific or general? However, the sample inefficiency of \emph{tabula rasa} deep reinforcement learning and the challenge of designing a single robot which can perform a wide variety of tasks has led to the current dominant paradigm of `one robot one task'. 
In stark contrast, domains like vision \citep{girshick2014rich, he2020momentum} and language \citep{dai2015semi, radford2018improving}, which are not plagued by the challenges of physical embodiment, have witnessed tremendous progress especially by leveraging large scale pre-training followed by transfer learning to many tasks through limited task-specific fine-tuning. Moreover, multiple domains are witnessing a confluence, with domain specific architectures being replaced by Transformers \citep{vaswani2017attention}, a general-purpose architecture with no domain-specific inductive biases. 

How can we bring to bear the advances in large-scale pre-training, transfer learning and general-purpose Transformer architectures, to the field of robotics? We believe that modular robot systems provide a natural opportunity by affording the flexibility of combining a small set of general-purpose building blocks into a task-optimized morphology. Indeed, modularity at the level of hardware is a motif which is extensively utilized by evolution in biological systems \citep{hartwell1999molecular, kashtan2005spontaneous} and by humans in many modern engineered systems. However, prior works \citep{wang2018nervenet, chen2018hardware, sanchez2018graph} on learning policies that can generalize across different robot morphologies have been limited to: (1) manually constructed variations of a single or few base morphologies, i.e. little diversity in the kinematic structure;  (2) low complexity of control ($\le 7$ degrees of freedom); (3) using Graph Neural Networks \citep{scarselli2008graph} based on the assumption that kinematic structure of the robot is the correct inductive bias. 

\begin{figure*}[t]
\centering
\includegraphics[width=0.9\linewidth]{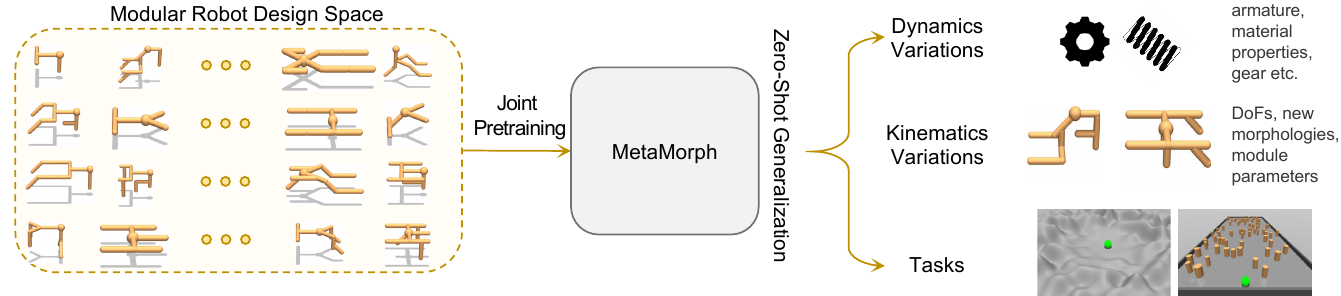}
  \caption{\textbf{Learning universal controllers.} Given a modular robot design space, our goal is to learn a controller policy, which can generalize to unseen variations in dynamics, kinematics, new morphologies and tasks. Video available at \href{https://metamorph-iclr.github.io/site/}{this project page}.}
  \vspace{\figvspace}
\label{fig:overview}
\end{figure*}

In this work, we take a step towards a more challenging setting (Fig.~\ref{fig:overview}) of learning a universal controller for a modular robot design space which has the following properties: (a) \textbf{generalization} to unseen variations in dynamics (e.g. joint damping, armature, module mass) and kinematics (e.g. degree of freedom, morphology, module shape parameters) and  (b) \textbf{sample-efficient} transfer to new morphologies and tasks. We instantiate the exploration of this general setting in the UNIMAL design space introduced by \cite{gupta2021embodied}. We choose the UNIMAL design space as it contains a challenging ($15-20$ DoFs) distribution of robots that can learn locomotion and mobile manipulation in complex stochastic environments. Learning a single universal controller for a huge variety of robot morphologies is difficult due to: (1) differences in action space, sensory input, morphology, dynamics, etc. (2) given a modular design space, not all robots are equally adept at learning a task, e.g. some robots might inherently be less sample-efficient \citep{gupta2021embodied}. 

To this end, we propose \maple, a method to learn a universal controller for a modular robot design space. \maple is based on the insight that robot morphology is just another modality on which we can condition the output of a Transformer. \maple tackles the challenge of differences in embodiment by leveraging a Transformer based architecture which takes as input a sequence of tokens corresponding to the number of modules in the robot. Each input token is created by combining proprioceptive and morphology information at the level of constituent modules. The combination of proprioceptive and embodiment modalities and large scale joint pre-training leads to policies which exhibit zero-shot generalization to unseen variations in dynamics and kinematics parameters and sample-efficient transfer to new morphologies and tasks. Finally, to tackle the differences in learning speeds of different robots, we propose \textit{dynamic replay buffer balancing} to dynamically balance the amount of experience collection for a robot based on its performance.

In sum, our key contributions are: (1) we introduce \maple to learn a universal controller for a modular design space consisting of robots with high control complexity for challenging 3D locomotion tasks in stochastic environments; (2) we showcase that our learned policy is able to zero-shot generalize to unseen variations in dynamics, kinematics, new morphologies and tasks, which is particularly useful in real-world settings where controllers need to be robust to hardware failures; (3) we analyze the learned attention mask and discover the emergence of motor synergies \citep{bernstein1966co}, which partially explains how \maple is able to control a large number of robots.

\section{Related Work}
Prior works on learning control policies which can generalize across robot morphologies have primarily focused on parametric variations of a single \citep{chen2018hardware} or few ($2-3$) robot types \citep{wang2018nervenet, sanchez2018graph, huang2020one, kurin2021my}. For generalizing across parametric variations of a single morphology, various approaches have been proposed like using a learned hardware embedding \citep{chen2018hardware}, meta-learning for policy adaptation \citep{al2017continuous, ghadirzadeh2021bayesian}, kinematics randomization \citep{exarchos2020policy}, and dynamics randomization \citep{peng2018sim}. In case of multiple different morphologies, one approach to tackle the challenge of differences in action and state spaces is to leverage Graph Neural Networks \citep{scarselli2008graph, Kipf2017SemiSupervisedCW, battaglia2018relational}. \citet{wang2018nervenet, huang2020one} use GNNs to learn joint controllers for planar agents ($\leq$ 7 DoFs). \cite{blake2021snowflake} propose freezing selected parts of networks to enable training GNNs for a single morphology but with higher control complexity. The usage of GNNs is based on the assumption that the robot morphology is a good inductive bias to incorporate into neural controllers, which can be naturally modelled by GNNs. Recently, \citet{kurin2021my} also proposed using Transformers for training planar agents. Our work differs from \citet{kurin2021my} in the diversity and scale of training robots, complexity of the environments, conditioning the Transformer on morphological information, and showcasing strong generalization to unseen morphologies and tasks (see  \S~\ref{appendix:baseline}).

Another closely related line of work is the design of modular robot design spaces and developing algorithms for co-optimizing morphology and control \citep{sims1994evolving} within a design space to find task-optimized combinations of controller and robot morphology. When the control complexity is low, evolutionary strategies have been successfully applied to find diverse morphologies in expressive soft robot design spaces \citep{cheney2014unshackle, cheney2018scalable}. In the case of rigid bodies, \citet{ha2019reinforcement, schaff2019jointly, liao2019data} have proposed using RL for finding optimal module parameters of fixed hand-designed morphology for rigid body robots. For more expressive design spaces, GNNs have been leveraged to share controller parameters \citep{wang2018neural} across generations or develop novel heuristic search methods for efficient exploration of the design space \citep{zhao2020robogrammar}. In contrast to task specific morphology optimization, \cite{joseph2021taskagnostic} propose evolving morphologies without any task or reward specification. Finally, for self reconfigurable modular robots \citep{fukuda1988dynamically, yim2007modular}, modular control has been utilized in both real \citep{rubenstein2014programmable, mathews2017mergeable} and simulated \citep{pathak2019learning} systems. 

\section{Learning a Universal Controller}
We begin by reviewing the UNIMAL design space and formulating the problem of learning a universal controller for a modular robot design space as a multi-task reinforcement learning problem.

\subsection{The UNIMAL Design Space}
An agent morphology can be naturally represented as a kinematic tree, or a directed acyclic graph, corresponding to a hierarchy of articulated 3D rigid parts connected via motor actuated hinge joints. The graph $\mathcal{G} := (\mathcal{V}, \mathcal{E})$ consists of vertices  $\mathcal{V} = \{v_1, ..., v_n\}$ corresponding to modules of the design space, and edges $e_{ij} \in \mathcal{E}$ corresponding to joints between $v_i$ and $v_j$. Concretely, in the UNIMAL \citep{gupta2021embodied} design space, each node represents a component which can be one of two types: (1) a sphere parameterized by radius and density to represent the head of the agent and form the root of the tree; (2) cylinders parameterized by length, radius, and density to represent the limbs of the robot. Two nodes of the graph can be connected via at most two motor-actuated hinge joints (i.e. $\mathcal{G}$ is a multi-graph), parameterized by joint axis, joint limits and a motor gear ratio.

\subsection{Joint Policy Optimization}
The problem of learning a universal controller for a set of $K$ robots drawn from a modular robot design space is a multi-task RL problem. Specifically, the control problem for each robot is an infinite-horizon discounted Markov decision process (MDP) represented by a tuple $(S, A, T, R, H, \gamma)$, where $S$ represents the set of states, $A$ represents the set of available actions, $T(s_{t+1}|s_t, a_t)$ represents the transition dynamics, $R(s, a)$ is a reward function, $H$ is the horizon and $\gamma$ is the discount factor. At each time step, the robot $k$ receives an observation $s_t^k$, takes an action $a_t^k$, and is given a reward $r_t^k$. A policy $\pi_\theta(a_t^k| s_t^k)$ models the conditional distribution over action $a_t^k \in A$ given state $s_t^k \in S$.  The goal is to find policy parameters $\theta$ which maximize the average expected return across all tasks: $R=\frac{1}{K}\sum_{k=0}^K\sum_{t=0}^H \gamma^t r_t^k$. We use Proximal Policy Optimization (PPO) \citep{schulman2017proximal}, a popular policy gradient \citep{williams1992simple} method for optimizing this objective. 

\section{\maple}
Progress in model-free reinforcement learning algorithms has made it possible to train locomotion policies for complex high-dimensional agents from scratch, albeit with tedious hyperparamter tuning. However, this approach is not suitable for modular design spaces containing exponentially many robot morphologies. Indeed, \citet{gupta2021embodied} estimates that the UNIMAL design space contains more than $10^{18}$ robots. Hence, learning a separate policy for each robot is infeasible. However, the modular nature of the design space implies that while each robot morphology is unique, it is still constructed from the same set of modules and potentially shares subgraphs of the kinematic tree with other morphologies. We describe how \maple exploits this structure to meet the challenge of learning a universal controller for different morphologies.

\subsection{Fusing Proprioceptive States and Morphology Representations}
To learn policies that can generalize across morphologies, we must encode not only proprioceptive states essential for controlling a single robot, but also morphological information.  From a multi-task RL perspective, this information can be viewed as a task identifier, where each task corresponds to a different robot morphology, all drawn from the same modular design space. Hence, instead of learning a policy which is \textit{agnostic} to the robot morphology, we need to learn a policy \textit{conditioned} on the robot morphology. Consequently, at each time step $t$ (we drop the time subscript for brevity), the robot receives an observation $s^k = (s^k_m, s^k_p, s^k_g)$ which is composed of the morphology representation ($s^k_m$), the proprioceptive states ($s^k_p$), and additional global sensory information ($s^k_g$). See \S~\ref{appendix:input_obs} for a detailed description of each observation type.

\subsection{Morphology Aware Transformer} \label{mat}

\begin{figure*}[t]
\centering
\includegraphics[width=0.96\linewidth]{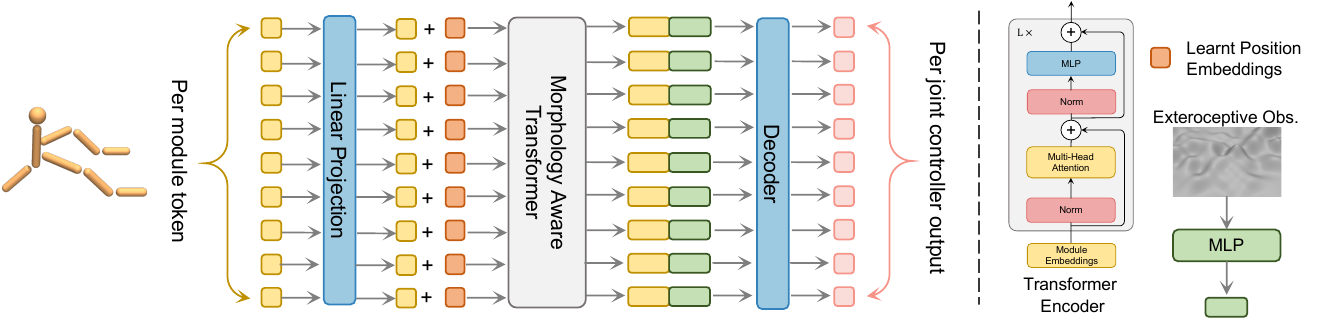}
  \caption{\textbf{\maple.} We first \textit{process} an arbitrary robot by creating a 1D sequence of tokens corresponding to depth first traversal of its kinematic tree. We then linearly embed each token which consists of proprioceptive and morphological information, add learned position embeddings and \textit{encode} the resulting sequence of vectors via a Transformer encoder. The output of the Transformer is concatenated with a linear embedding of exteroceptive observation before passing them through a \textit{decoder} to output per joint controller outputs.}
  \vspace{\figvspace}
\label{fig:system}
\end{figure*}

The robot chooses its action via a stochastic policy $\pi_\theta(a_t^k| s_t^k)$ where $\theta$ are the parameters of a pair of deep neural networks: a policy network that produces an action distribution (Fig.~\ref{fig:system}), and a critic network that predicts discounted future returns. We use Transformers \citep{vaswani2017attention} to parametrize both policy and critic networks as described in detail below. 

\textbf{Encode.} We make a distinction between how we process local and global state information. Concretely, let $s^k = (s^k_l, s^k_g)$ where $s^k_l = (s^k_m, s^k_p)$. Since the number of joints between two modules can vary, we zero pad $s^k_{l_i}$ to ensure that input observation vectors are of the same size, i.e. $s^k_l \in \mathbb{R}^{N \times M}$. In order to provide an arbitrary robot morphology as input to the Transformer, we first create a 1D sequence of local observation vectors  by traversing the kinematic tree in depth first order starting at the root node (torso in case of the UNIMAL design space). We then apply a single layer MLP independently to $s^k_{l_i}$ to create a $D$ dimensional module embedding. We also add learnable 1D position embeddings to the module embeddings to automatically learn positional information:
\begin{align}
    \mbf{m}_0 &= [\phi(s^k_{l_1}; \mbf{W}_e); \cdots; \,\phi(s^k_{l_{N}}; \mbf{W}_e)] + \mbf{W}_{\text{pos}},
\label{eq:embedding}
\end{align}

where $\phi(\cdot)$ is the embedding function, $\mbf{W}_e \in \mathbb{R}^{M \times D}$ are the embedding weights, and $\mbf{W}_\text{pos} \in \mathbb{R}^{N \times D}$ are the learned positional embeddings. Note that in practice we zero pad the input sequence of local observation vectors for efficient batch processing of multiple morphologies.

\textbf{Process.} From the module embeddings described above, we obtain the output feature vectors as:
\begin{align}
    \mbf{m^\prime}_\ell &= \op{MSA}(\op{LN}(\mbf{m}_{\ell-1})) + \mbf{m}_{\ell-1}, && \ell=1\ldots L \label{eq:msa_apply} \\
    \mbf{m}_\ell &= \op{MLP}(\op{LN}(\mbf{m^\prime}_{\ell})) + \mbf{m^\prime}_{\ell}, && \ell=1\ldots L  \label{eq:mlp_apply}
\end{align}
where MSA is multi-head attention \citep{vaswani2017attention} and LN is Layernorm \citep{ba2016layer}.

\textbf{Decode.} We integrate the global state information $s^k_g$ consisting of high-dimensional sensory input from camera or depth sensors. Naively concatenating $s^k_g$ and $s^k_l$ in the encoder has two downsides: (1) it dilutes the importance of low-dimensional local sensory and morphological information; (2) it increases the number of Transformer parameters due to an increase in the dimensionality of the input embedding ($D$). Instead, we obtain the outputs of policy network as follows:
\begin{align}
    \mbf{g} = \gamma(s^k_g; \mbf{W}_g), 
    && \mbf{\mu(s^k)} = \phi(\mbf{m_{l_i}}, \mbf{g}; \mbf{W}_d), 
    && \pi_\theta(a^k| s^k) = \mathcal{N}( \{\mu(s^k_i)\}^{N}_i, \Sigma),
\end{align}
where $\gamma(\cdot)$ is a 2-layer MLP with parameters $\mbf{W}_g$, $\phi(\cdot)$ is an embedding function with $\mbf{W}_d$ as the embedding weights. The action distribution is modeled as a Gaussian distribution with a state-dependent mean $\mu(s^k_i)$  and a fixed diagonal covariance matrix $\Sigma$. Similarly, for the critic network, we estimate value for the whole morphology by averaging the value per limb.

\subsection{Dynamic Replay Buffer Balancing} \label{drb}
Joint training of diverse morphologies is challenging as different morphologies are adept at performing different tasks. Consequently, some morphologies might be inherently better suited for the pre-training task. Let us consider two robots: (A) Robot A locomotes in a falling forward manner, i.e., robot A is not passively stable; (B) Robot B is passively stable. Especially with early termination, robot A will keep falling during the initial phases of training, which results in shorter episode lengths, whereas robot B will have longer episode lengths. Hence, more data will be collected for robot B in the earlier phases of training, and in turn will lead to robot B learning even faster, thereby resulting in a `\textit{rich gets richer}' training dynamic. However, our goal is to ensure that all morphologies have a similar level of performance at the end of training as we want to generalize across the entire distribution of morphologies. 

To address this issue, we propose a simple \textit{dynamic replay buffer balancing} scheme. On-policy algorithms like PPO \citep{schulman2017proximal} proceed in iterations that alternate between experience collection and policy parameter update. Let $\mathcal{E}_k$ be any performance metric of choice, e.g. normalized reward, episode length, success ratio, etc. Let $\tau$ be the training iteration number. At each iteration, we sample the $k^{\text{th}}$ robot with probability $P_k$, given by:
\begin{align}
    P_k = \frac{\mathcal{E}_k^{\beta}}{\sum \mathcal{E}_i^{\beta}}
    && \mathcal{E}_k^{\tau} = \alpha \mathcal{E}_k^{\tau} + (1 - \alpha) \mathcal{E}_k^{(\tau - 1)}, \label{eq:rb}
\end{align}
where $\alpha \in [0, 1]$ is the discount factor and the exponent $\beta$ determines the degree of dynamic prioritization, with $\beta = 0$ corresponding to the uniform case. In practice, we use episode length as our performance metric. We determine $P_k$ by replacing $\mathcal{E}_i$ with $\frac{1000}{\mathcal{E}_i}$, where the numerator is the maximum episode length.
\section{Experiments}
\begin{figure*}[t]
\centering
\includegraphics[width=0.85\linewidth]{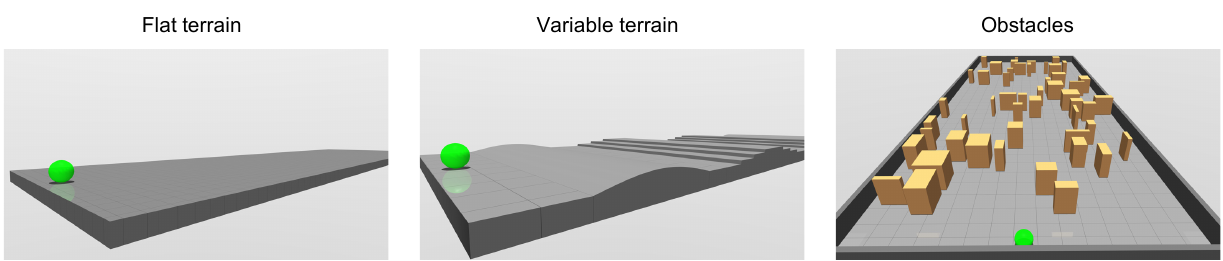}
  \caption{\textbf{Environments.} We evaluate our method on $3$ locomotion tasks: Flat terrain, Variable terrain: consists of three stochastically generated obstacles: hills, steps and rubble, and Obstacles: cuboid shaped obstacles of varying sizes.}
\label{fig:envs}
\end{figure*}
In this section, we evaluate our method \maple in different environments, perform extensive ablation studies of different design choices, test zero-shot generalization to variations in dynamics and kinematics parameters, and demonstrate sample efficient transfer to new morphologies and tasks. For qualitative results, please refer to the video on our project website \footnote{\url{https://metamorph-iclr.github.io/site/}}.

\subsection{Experimental Setup}
We create a training set of $100$ robots from the UNIMAL design space \citep{gupta2021embodied} (see \S~\ref{appendix:env}). We evaluate \maple on three different environments (Fig.~\ref{fig:envs}) using the MuJoCo simulator \citep{todorov2012mujoco}. In all $3$ environments the goal of the agent is to maximize forward displacement over the course of an episode which lasts $1000$ timesteps. The $3$ environments are: (1) Flat terrain (FT); (2) Variable terrain (VT): VT is an extremely challenging environment as during each episode a new terrain is generated by randomly sampling a sequence of terrain types and interleaving them with flat terrain. We consider $3$ types of terrains: hills, steps, and rubble; (3) Obstacles: cuboid shaped obstacles of varying sizes on flat terrain.

We use a dense morphology \textit{independent} reward function for all our tasks as it is not feasible to design a reward function tailored to each morphology. In all tasks, our reward function promotes forward movement using small joint torques (the latter obtained via a small energy usage penalty).  In addition, as described in \S\ref{drb}, we use early termination across all environments when we detect a fall (i.e. if the torso height drops below $50\%$ (FT, Obstacles) or $30\%$ (VT) of its initial height).

\begin{figure*}[t]
\centering
\includegraphics[width=1.0\linewidth]{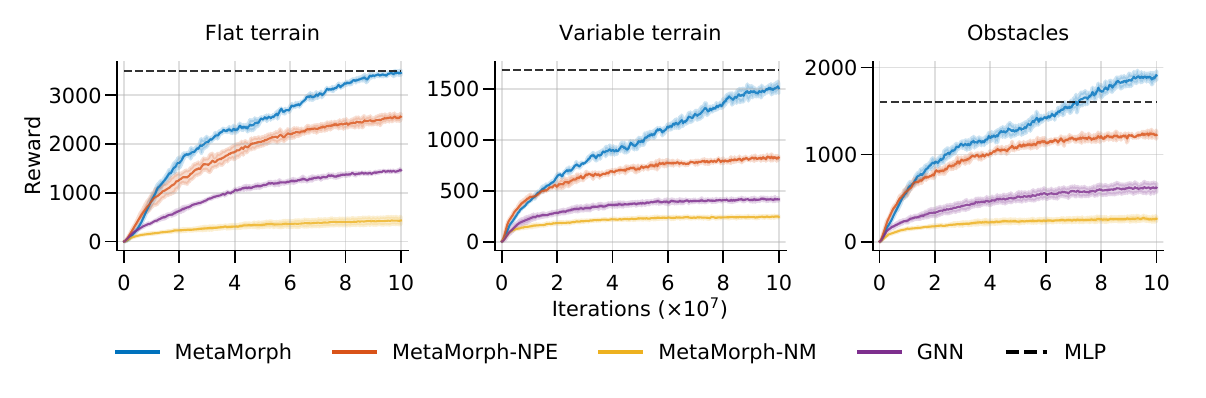}
  \caption{\textbf{Joint pre-training of modular robots.} Mean reward progression of $100$ robots from the UNIMAL design space averaged over $3$ runs in different environments for baselines and ablations described in \S~\ref{section:baseline}. Shaded regions denote standard deviation. Across all $3$ environments, \maple consistently outperforms GNN, and is able to match the average reward achieved by per morphology MLP baseline, an approximate upper bound of performance.}
\label{fig:baselines}
\end{figure*}

\begin{wrapfigure}[18]{r}{0.4\textwidth}
  \vspace{-0.3in}
  \begin{center}
    \includegraphics[width=0.4\textwidth]{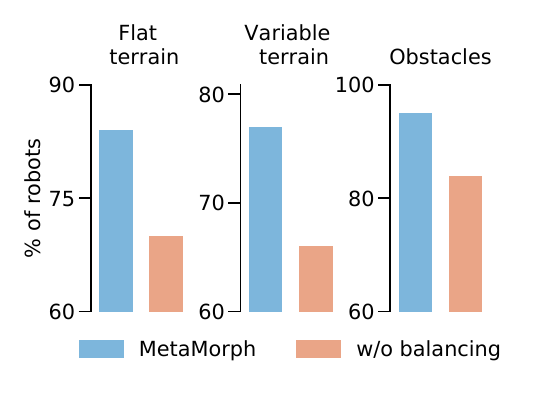}
  \end{center}
  \vspace{-0.3in}
    \caption{\textbf{Importance of dynamic replay buffer balancing.} We compare the percentage of robots with final performance greater than $75\%$ of the MLP baseline performance when trained jointly. Across all $3$ environments, on average for $10-15\%$ robots, \maple w/o balancing is unable to learn a good control policy.}
    \label{fig:buffer}
\end{wrapfigure}

\subsection{Baselines and Ablations}\label{section:baseline}
\textbf{Baselines}: We compare against the following baselines:

(1) \emph{GNN}: We modify the NerveNet model proposed by \citet{wang2018nervenet} to learn control policies for complex 3D morphologies with variable number of joints. Specifically, we replace how the NerveNet model receives input and produces output by our encode and decode steps respectively (\S\ref{mat}). In addition, the model receives the same observation as MetaMorph i.e. $s^k = (s^k_m, s^k_p, s^k_g)$ and is trained with dynamic replay buffer sampling. Thus, the only difference is in the \textbf{process} step. This helps test if the domain-specific inductive bias of the robot kinematic tree in the GNN is necessary. 

(2) \emph{MLP}: We train all $100$ robots separately with a 2-layer MLP and report the average performance. This baseline serves as an approximate upper bound for our method.

\textbf{Ablations}: We also do an ablation study of different design choices involved in our method. We refer our full method as \maple and consider the following ablations: (1) \maple-NPE: no learned position embeddings; (2) \maple-NM: we only provide $s^k = (s^k_p, s^k_g)$ as inputs, i.e. the model does not have access to information about the robot morphology. 

Fig.~\ref{fig:baselines} shows learning curves across $3$ environments for training $100$ morphologies.  In all environments \maple can successfully match the average reward achieved via the per morphology MLP baseline on both FT, and obstacle environment. While \maple performance in VT is slightly below the MLP baseline, we note that it has not saturated and we stopped training at $10^8$ iterations across all three environments. Moreover, \maple is significantly more sample-efficient ($5\times$) than training independent MLPs ($5\times10^6$ iterations per robot). 
\begin{figure*}[t]
\centering
\includegraphics[width=0.9\linewidth]{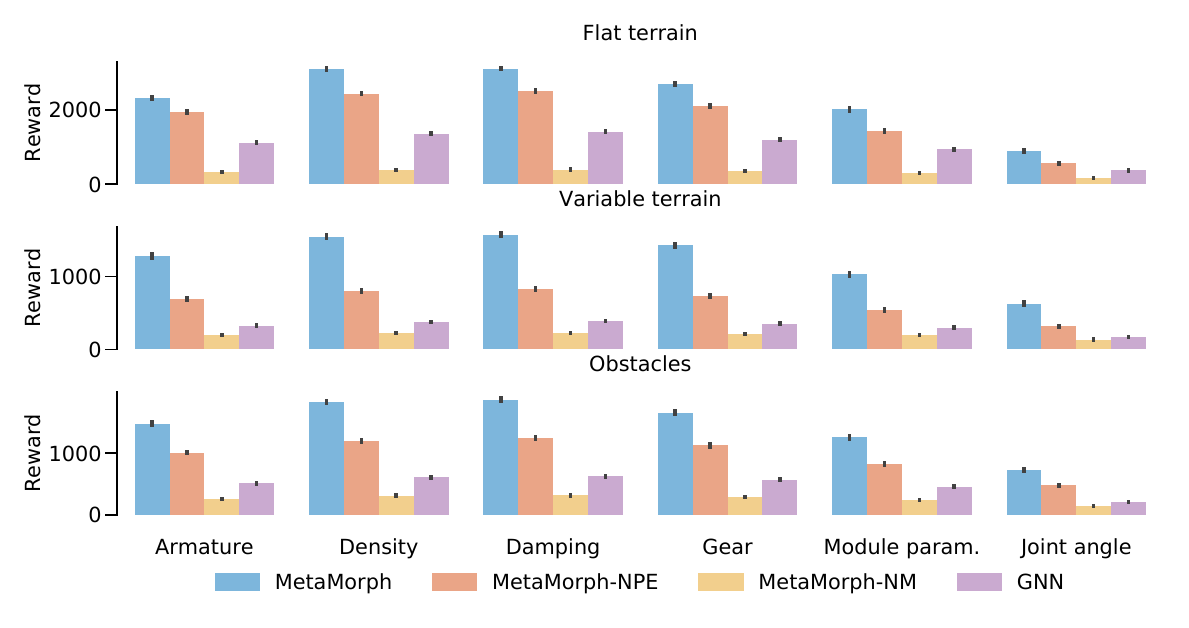}
  \caption{\textbf{Zero-shot generalization.} We create $400$ new robots for each type of variation in dynamics parameters (armature, density, damping, gear) and kinematics parameters (module shape, joint angle). Bars indicate average zero-shot reward over $10$ trials for $400$ robots and error bars denote $95\%$ bootstrapped confidence interval. Across all types of variations, and environments we find that \maple considerably outperforms GNN, and \maple-NM.}
  \vspace{\figvspace}
\label{fig:dk}
\end{figure*}
The GNN baseline saturates at a level $2$ to $3$ times below \maple. In GNN based models, locality and neighborhood connectivity is explicitly baked into the model. Interestingly, just like ViT \citep{dosovitskiy2021an} sparingly utilizes the 2D neighborhood structure of images at the beginning of the model by cutting the image into patches, \maple uses the graph structure of the robot in the beginning by creating a 1D sequence corresponding to the kinematic tree by traversing the graph in depth first order. Moreover, the position embeddings carry no information about the graph structure and are learned from scratch. We highlight that the learned position embeddings significantly improve the performance of \maple, just as they do in Transformer based image classification.  Finally, without access to the morphological information, \maple-NM fails to learn a policy that can control diverse robot morphologies. All of this substantiates our central claim that morphological state information is necessary to learn successful control policies, although the kinematic graph need not be explicitly baked into neural architectures to learn policies capable of controlling diverse robot morphologies. Finally, we test the importance of dynamic replay buffer balancing in Fig.~\ref{fig:buffer}, and find that balancing is necessary to learn a good control policy in $10-15\%$ of robots across all $3$ environments.

\subsection{Zero-Shot Generalization} \label{section:zero_shot_gen}
Our focus in this work is to learn policies that can generalize to unseen robots drawn from a modular robot design space. In this section, we demonstrate that \maple shows favorable generalization properties across many different kinematic and dynamic variations. 

\textbf{Experimental Setup.} For each of the $100$ training robots, we create a dedicated test set to test zero-shot transfer performance across two types of variations: dynamics (armature, density, damping, and motor gear) and kinematics (module shape parameters like radius, and length of cylindrical limbs, and joint angles). For each training robot, we randomly create $4$ different variants for each property, i.e. $400$ robots with armature variations, and so on.  While creating a new variant, we change the relevant property of all modules or joints. See Table~\ref{table:dyn_kin_var_parm} for sampling ranges. We then compare zero-shot performance averaged over $10$ trials.

\textbf{Generalization: Dynamics.} First, we consider generalization to different dynamics (Fig.~\ref{fig:dk}). We find consistently that \maple performs significantly better than \maple-NM and GNN across all types of dynamic variations and all environments. In fact, the difference is more pronounced for harder tasks like VT and Obstacles. We note that this result is surprising as we do not do dynamics randomization during training, i.e., all robots in the training set have the same armature and damping parameters. Despite this, we see strong generalization performance. 

\textbf{Generalization: Kinematics.} Next, we consider generalization to different kinematics parameters (Fig.~\ref{fig:dk}). This is a significantly more challenging setting as the model has to generalize to unseen variations in module shape parameters and changes to joint angle ranges. In fact, changes to joint angles can significantly alter the range of possible motion and might necessitate a different gait for successful locomotion. Consequently, we find that even though \maple exhibits strong generalization performance compared to \maple-NM and GNN in all $3$ environments, there is indeed a performance drop for the challenging setting of variations in joint angles. However, the zero-shot performance is encouraging and motivates our next set of experiments on transfer learning.
\begin{figure*}[t]
\centering
\includegraphics[width=0.9\linewidth]{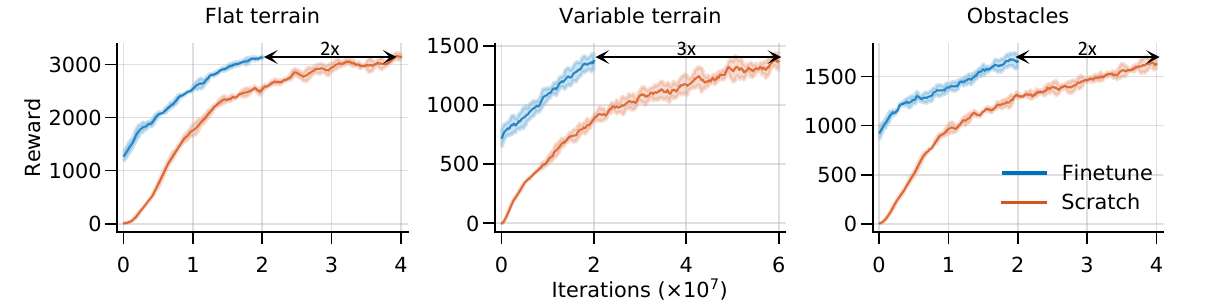}
  \caption{\textbf{Fine tuning: New robot morphologies.} Comparison of reward progression of $100$ new robot morphologies averaged over $3$ runs for pre-trained \maple in the same environment vs from scratch. Shaded regions denotes standard deviation. Across all environments pre-training leads to strong zero-shot performance and $2-3 \times$  more sample efficiency.}
\label{fig:gen_mor}
\end{figure*}

\begin{figure*}[t]
\centering
\includegraphics[width=0.9\linewidth]{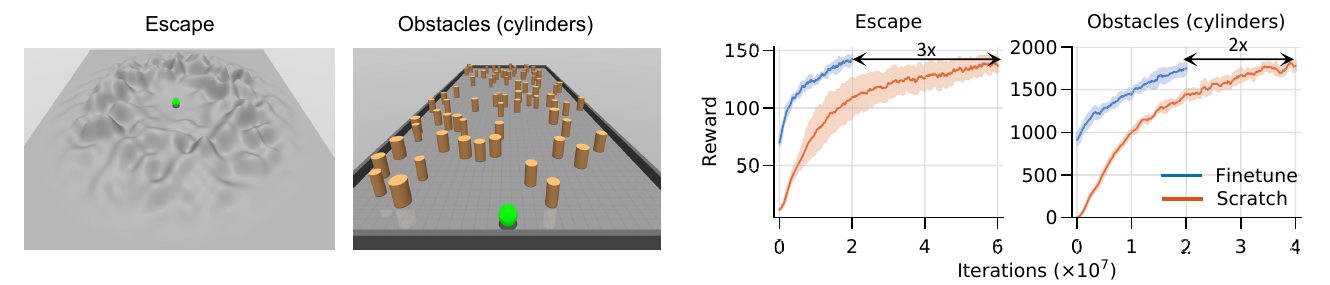}
  \caption{\textbf{Fine tuning: New robot morphologies and tasks.} \textit{Left}: Test environments. \textit{Right}: Comparison of reward progression of $100$ test robots averaged over $3$ runs for pre-trained \maple (VT $\rightarrow$ Escape, Obstacles $\rightarrow$ Obstacles (cylinders)) vs from scratch. Shaded regions denotes standard deviation. Across all environments pre-training leads to strong zero-shot performance and $2-3 \times$ savings in training iterations to achieve the same level of average reward.}
\label{fig:gen_env}
\end{figure*}

\subsection{Sample Efficient Transfer Learning}
\textbf{Experimental Setup.} Here we create $100$ robots from the UNIMAL design space which were not part of the training set, and we demonstrate that \maple shows favorable sample efficient transfer to unseen morphologies, and even unseen morphologies performing novel tasks.

\textbf{Different morphologies.} We first consider sample efficient learning of controllers for new morphologies on the same task by taking a pre-trained \maple model on an environment, and then fine tuning it to learn to control morphologies in the test set. In Fig.~\ref{fig:gen_mor} we compare the number of training iterations required to achieve a particular performance level when we fine tune \maple vs training from scratch on the test set. Across all $3$ environments we not only find strong zero-shot performance , but also $2$ to $3$ times higher sample efficiency compared to training from scratch.

\textbf{Different morphologies and novel tasks.} Finally, we consider the most realistic and general setting mimicking the promise of modular robots, where we are faced with a novel task and want to use a new robot morphology which may be suited for this task.  We consider the same set of $100$ test robots on two new tasks (Fig.~\ref{fig:gen_env}): (1) Escape: The agent starts at the center of a bowl shaped terrain surrounded by small hills (bumps), and has to maximize the geodesic distance from the start location (escape the hilly region). (2) Obstacles (cylinders): cylinder shaped obstacles of varying sizes (the size distribution is also different from the train task i.e. cuboid shapes). We transfer the learned policy from VT and Obstacles to Escape and Obstacles-Cylinders respectively. Again, we find that there is a strong zero-shot performance across all $3$ environments and fine-tuning is $2$ to $3$ times more sample efficient than training from scratch.

\subsection{Emergent Motor Synergies}
\begin{figure*}[t]
\vspace{-12mm}
\centering
\includegraphics[width=0.90\linewidth]{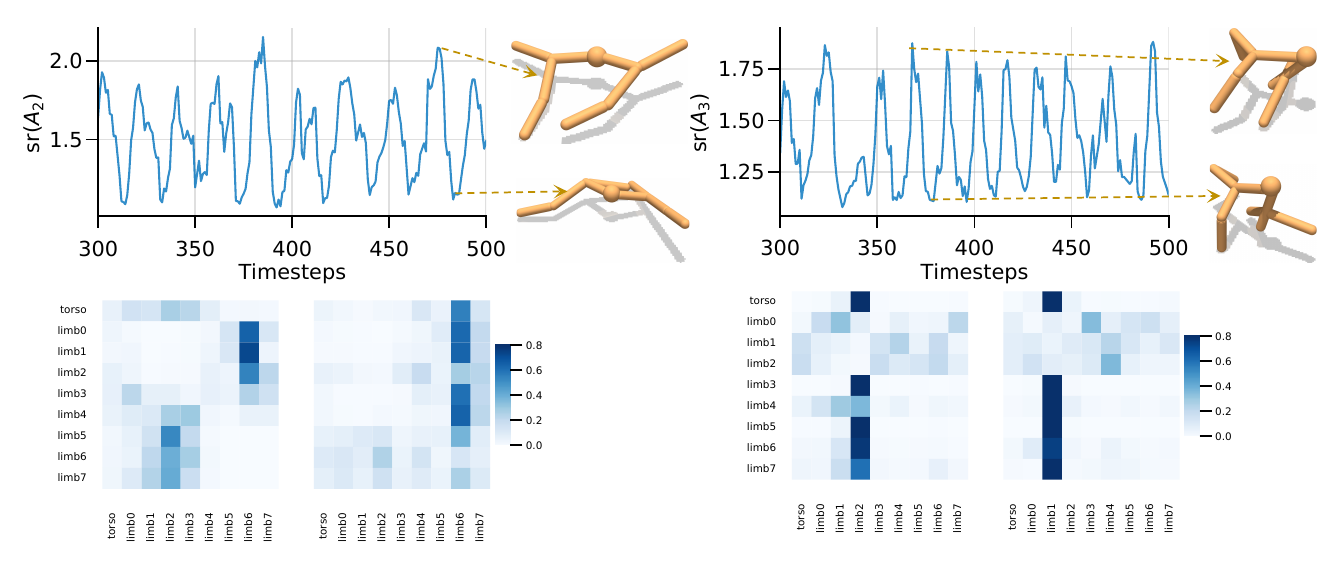}
  \caption{\textbf{Emergent motor synergies.} We plot the stable rank of the attention matrix ($\mbf{\text{sr}}(A_l)$ ) for two agents performing locomotion on a flat terrain. $\mbf{\text{sr}}(A_l)$ is small and oscillates between two values which correspond to attention maps where groups of limbs are activated simultaneously (denoted by dark columns), a characteristic signature of motor synergies.}
  \vspace{-5mm}
\label{fig:mot_syn}
\end{figure*}
We next search for a potential explanation for how \maple can coordinate the large number of DoFs ($\sim 16 \times 100$) across several agents. Our hypothesis is inspired by \citet{bernstein1966co}, which proposed the existence of muscle synergies as a neural strategy to simplify the control of multiple DoFs by the central nervous system. A muscle synergy corresponds to the constrained movements of sets of joints (or muscles) through co-activation by a single neural signal. Such synergies obviate the need to control all joints independently, by coupling sets of joints into adaptive functional groups.  

Although the definition of synergies \citep{bruton2018synergies} vary in the literature, dimensionality reduction is generally accepted as a signature of synergistic control \citep{todorov2004analysis, todorov2004optimality}. Consequently, to test this hypothesis, in Fig.~\ref{fig:mot_syn} we plot the stable rank of the attention matrix. For attention matrix $A_l \in \mathbb{R}^{m \times m}$ for layer $l$, the stable rank is defined as: $\mbf{\text{sr}}(A_l) = \frac{\lVert A_l \rVert^2_F}{\lVert A_l \rVert_2} = \frac{\sum \sigma_i^2}{\sigma_{max}^2}$, where $\sigma_i$ are the singular values of $A_l$. We note that $\mbf{\text{sr}}(A_l)$ is small and oscillates between two values which correspond to attention maps where groups of limbs are activated simultaneously (denoted by dark columns), a characteristic signature of motor synergies. Hence, \maple simplifies the control problem by learning to activate different motor synergies depending on both $s^k_m$ and $s^k_p$.

\section{Conclusion}
In this work, we explored how we can learn a universal controller for a large modular robot design space. To this end, we proposed \maple, a Transformer approach based on the insight that robot morphology is just another modality on which we can condition the output of a Transformer. We showcased that pre-training on a large collection of diverse morphologies leads to policies which can generalize to unseen variations in kinematics, dynamics, new morphologies, and tasks. We hope that our work serves as a step towards realizing the potential of large-scale pre-training and fine-tuning in the field of robotics, a paradigm that has seen tremendous success in vision and language. 

\section{Acknowledgement}
We gratefully acknowledge the support by Department of Navy awards (N00014-16-1-2127, N00014-19-1-2477) issued by the Office of Naval Research.

\section{Reproducibility Statement}
We have released a PyTorch \citep{pytorch} implementation of MetaMorph on GitHub (\url{https://github.com/agrimgupta92/metamorph}). In addition, the repository also includes all the robots and environments used for benchmarking.

\bibliography{references}
\bibliographystyle{conf}
\clearpage

\appendix
\section{Implementation Details} 
In this section, we provide additional implementation details for our method.

\subsection{Input Observations} \label{appendix:input_obs}
In our setup, at each time step $t$ (we drop the time subscript for brevity) the robot receives an observation $s^k = (s^k_m, s^k_p, s^k_g)$ which is composed of the morphology representation ($s^k_m$), the proprioceptive states ($s^k_p$) and additional global sensory information ($s^k_g$).  We note that prior work \citep{kurin2021my, huang2020one} has defined morphology as the connectivity of the limbs. However, connectivity is only one aspect of morphology, and in general a substantial amount of additional information may be required to adequately describe the morphology. Consider a modular robot composed of  $n \in {1...N_k}$ modules. For each module $s^k_{m_i}$ consists of: (1) \textit{Module Parameters}: Module shape parameters (e.g. radius, height), material information (e.g. density), and local geometric orientation and position of the child module with respect to the parent module. (2) \textit{Joint Parameters}: This consists of information about joint type (e.g. hinge) and its properties (e.g. joint range and axis), and actuator type (e.g. motor) and its properties (e.g. gear). All this information can be found for example, in the Universal Robot Description Format (URDF) or in simulator specific kinematic trees (e.g. MuJoCo XML \citep{todorov2012mujoco}). 

Similarly for each module $s^k_{p_i}$ consists of the instantaneous state of the system: (1) \textit{Module Proprioception}: 3D Cartesian position, 4D quaternion orientation, 3D linear velocity, and 3D angular velocity. (2) \textit{Joint Proprioception}: Position and velocity in the generalized coordinates. Except the root node, each module will be connected to its parent module via a set of joints. We adopt the convention of providing joint parameters and proprioception information to the child node. We note that the importance of proprioceptive observations has also been observed in learning good locomotion \citep{lee2020learning, kumar2021rma} and manipulation policies \citep{mandlekar2021what}.

In general, additional global sensory information ($s^k_g$) can be camera or depth sensor images. To save computation, we provide the information about the terrain as 2D heightmap sampled on a non-uniform grid. The grid is created by decreasing the sampling density as the distance from the root of the body increases. All heights are expressed relative to the height of the ground immediately under the root of the agent. The sampling points range from $1m$ behind the agent to $4m$ ahead of it along the direction of motion, as well as $4m$ to the left and right.

\subsection{Environments} \label{appendix:env}
All our training and testing environments are implemented in MuJoCo \citep{todorov2012mujoco}. For a detailed description about the environments and reward functions please refer \citet{gupta2021embodied}. \citet{gupta2021embodied} evolved UNIMALS in $3$ different environments: flat terrain, variable terrain and manipulation in variable terrain. For each environment at the end of evolution, they had $100$ task optimized robots. Out of these $300$, we choose a subset of $100$ UNIMALS as our train set. We ensure that no robot has the exact same kinematic tree or kinematic parameters. Similarly, we created a test set of $100$ UNIMALS. Finally, for zero shot evaluation experiments we created the variants as described in \S~\ref{section:zero_shot_gen}. For creating kinematic variations for joint angles, we randomly selected a joint range for each joint from Table~\ref{table:dyn_kin_var_parm} which had alteast $50\%$ overlap with the original joint angle range. This helped in preventing robot variants which could not be trained.

\subsection{Training Hyperparameters} \label{appendix:train_details}
We use Transformers \citep{vaswani2017attention} to parametrize both policy and critic networks. Global sensory information is encoded using a 2-layer MLP with hidden dimensions $[64, 64]$. We use Proximal Policy Optimization \citep{schulman2017proximal} for joint training of agents in all environments. All hyperparameters for Transformer and PPO are listed in Table~\ref{table:ppo_hyper}. In addition, we use dynamic replay buffer sampling with $\alpha = 0.1$ and $\beta = 1.0$. For all our ablations and baselines, we use the same hyperparameters. For MLP baseline we use a 2-layer MLP with hidden dimensions $[64, 64]$ and for GNN baseline we use a 5-layer MLP with hidden dimension $512$. The details of the joint training of modular robots are shown in Algorithm~\ref{alg:maple}. We ensure that all baselines and ablations have approximately the same number of parameters ($\sim 3.3$ Million). 

\subsection{Evaluation Methodology} \label{appendix:eval_method}
In general, the performance of RL algorithms is known to be strongly dependent on the choice of seed for random number generators \citep{henderson2018deep}. Hence, we run all our baseline and ablation experiments for $3$ random seeds. We found that the training curves were robust to the choice of seeds as indicated by small standard deviation (Fig.~\ref{fig:baselines}, ~\ref{fig:appendix_ablations}). Consequently, we evaluate the zero shot performance using pre-trained model corresponding to a single seed (Fig.~\ref{fig:dk}). For our transfer learning experiments (Fig.~\ref{fig:gen_mor}, ~\ref{fig:gen_env}), we fine tune the model with the random seed corresponding to the one used for pre-training (for all $3$ seeds).

\section{Additional Experiments}

\subsection{Baselines and Ablations} \label{appendix:baseline}
\textbf{Baselines}: We compare against the following baselines:

(1) \emph{\maple-AO}: Direct comparison to Amorpheus \citep{kurin2021my} is not feasible due to the following reasons: (1) does not work with 3D morphologies with variable number of joints between limbs; (2) does not incorporate exteroceptive observations; 
(3) Amorpheus ensures a balanced collection of experience by sequentially collecting data on each morphology and maintaining a separate replay buffer per morphology. Further, updates to policy parameters are also performed sequentially. This sequential nature of the algorithm is not amenable to scaling, and would require $\sim 30$ GPU days to train for 100 million iterations on Nvidia RTX 2080, while our method only needs 1.5 GPU days ($\sim$ \textbf{20x training speedup}).  

Hence, we compare with \maple-AO (\underline{\textbf{A}}morpheus \underline{\textbf{O}}bservation) as the closest variant to Amorpheus, where we provide the same input observations as described in \cite{kurin2021my}. Specifically, we provide the following input observations: one hot encoding of unique limb ID, 3D Cartesian position, 4D quaternion orientation, 3D linear velocity, and 3D angular velocity, position in generalized coordinates, and normalized joint angle ranges. Note that although both \maple and Amorpheus don't explicitly incorporate the graph structure as input to the policy, our input observations include additional morphology information (see \S~\ref{appendix:baseline}). In contrast, except for joint angle ranges, Amorpheus does not incorporate morphology information. 

\begin{wrapfigure}[17]{r}{0.6\textwidth}
  \vspace{-0.3in}
  \begin{center}
    \includegraphics[width=0.6\textwidth]{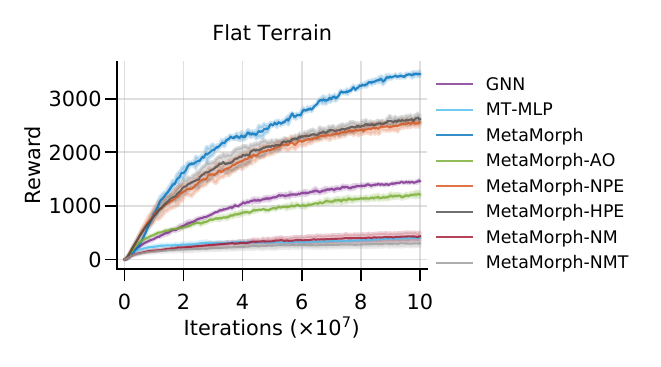}
  \end{center}
  \vspace{-0.3in}
      \caption{\textbf{Baselines and ablations.} Mean reward progression of $100$ robots from the UNIMAL design space averaged over $3$ runs in flat terrain for baselines and ablations described in \S~\ref{section:baseline} and \S~\ref{appendix:baseline}. Shaded regions denote standard deviation.}
    \label{fig:appendix_ablations}
\end{wrapfigure}

(2) \emph{Multi-Task MLP}: We train a 6-layer MLP with a hidden dimension of $512$. The MLP receives the same observations as \maple, and has the same number of parameters.

\textbf{Ablations}: We perform additional ablation studies to understand the importance of learnt position embeddings and morphology information in input observation. We refer our full method as \maple and consider the following additional ablations: (1) \maple-NMT (\underline{\textbf{N}}o  \underline{\textbf{M}}orphology information + \underline{\textbf{T}}ask encoding): we only provide $s^k = (s^k_p, s^k_g)$ as inputs, i.e. the model does not have access to information about the robot morphology. In addition, we provide a context token with binary encoding to identify the morphology; (2) \maple-HPE (\underline{\textbf{H}}and-designed \underline{\textbf{P}}osition \underline{\textbf{E}}ncoding): we provide  $s^k = (s^k_m, s^k_p, s^k_g)$  as inputs, with $s^k_m$ containing a per limb unique ID.

Fig.~\ref{fig:appendix_ablations} shows the learning curves for joint training in the flat terrain environment for $100$ training morphologies. Multi-task MLP baseline struggles to learn a good policy, which we attribute to the difficulty of the task due to the rich diversity of morphologies. We next investigate the role of learnt position embeddings and morphological information. \maple-HPE performs slightly better than \maple-NPE, indicating that adding unique limb ids improves performance. However, there is still a large gap between \maple and \maple-HPE, suggesting that learnt position embeddings capture additional information. Although \maple-AO performs better than \maple-NM due to the addition of unique limb ids and joint angle ranges, it is still significantly worse than \maple as it does not incorporate morphological information. Finally, \maple-NMT also performs poorly due to the lack of morphological information. 

\subsection{Position Embedding} \label{appendix:pos_embed}
\begin{figure*}[t]
\centering
\includegraphics[width=1.0\linewidth]{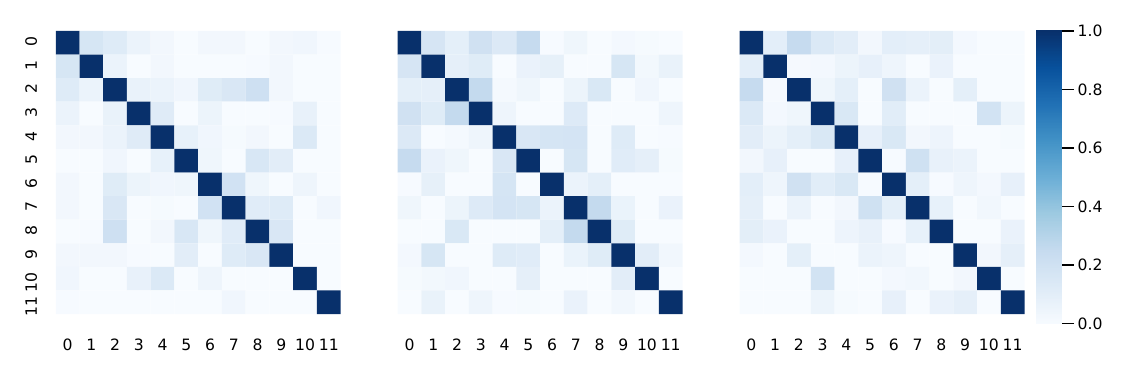}
  \caption{\textbf{Position embeddings.} Cosine similarity of position embeddings of \maple for $3$ different seeds in flat terrain environment.}
\label{fig:pos_embed}
\end{figure*}

To better understand the information captured by position embeddings we visualize their cosine similarity. We found that across all seeds and in all environments (we only show flat terrain for brevity) a diagonal pattern emerged (Fig~\ref{fig:pos_embed}). Although, this might suggest that learnt position embeddings are capturing a unique position identifier, we found that manually specifying a unique limb identifier to perform much worse (see \maple-HPE in Fig.~\ref{fig:appendix_ablations}). Thus, indicating that these embeddings are capturing additional useful information. We believe that further investigation is needed to better understand the information captured by position embeddings and defer it to future work.

\subsection{Kinematic tree traversal order} \label{appendix:dfs}
We first \textit{process} an arbitrary robot by creating a 1D sequence of tokens corresponding to the depth first traversal of its kinematic tree. We note that the DFS ordering is not unique, and for nodes with similar depth, we adopted the convention of visiting the nodes which come first in the MuJoCo XML. This convention is also adopted by MuJoCo when parsing the XML. We found that zero-shot transfer of the learnt policy was not robust to an opposite ordering of nodes and the performance dropped by $\sim 75\%$. However, we found that the policy could be made robust to variations in the DFS ordering by training the model with a simple data augmentation strategy of randomizing the visitation order of nodes at the same depth. 

\section{Limitations And Future Work}
In this work, we explored how we can learn a universal controller for a large modular robot design space. We make a key assumption that we already had access to a large collection of robots which were optimized for the task of locomotion. Hence, currently we require a two stage pipeline where we first create robots and then pre-train them jointly. An important direction of future work would be to combine these two phases in an efficient manner. Moreover, although we found that \maple has strong zero-shot generalization to unseen variations in kinematics and dynamics, there is indeed a performance drop on zero-shot generalization to new morphologies. Hence, an important avenue of exploration is creating algorithms for sample efficient transfer learning. Finally, our current suite of tasks focuses primarily on locomotion skills. An important line of future work will involve designing general purpose controllers which could perform multiple skills.

\begin{algorithm}[tb]
\caption{\maple: Joint Training of Modular Robots}
\label{alg:maple}
\begin{algorithmic}[1]
  \STATE {\bfseries Input:}\\
  $\pi_{\theta}$: policy function\\
  $V_{\phi}$: value function\\
  $R$: replay buffer \\
  $K$: training pool of robots \\
  $P_k$: probability of sampling robot $k$\\
  \hrulefill
  \STATE {\bfseries Initialize:} Learnable parameters $\theta$ for $\pi_\theta$, $\phi$ for $V_\phi$, $P$ uniform distribution
  \FOR {$i$=1,2,..$N_\text{iter}$}
  \STATE \codecomment{\# Collect robot experience in parallel}
  \FOR {$j$=1,2,..$N_\text{workers}$}
  \STATE Sample a robot $k \in K$, according to \Eqref{eq:rb}
  \STATE \codecomment{\# Collect one episode of experience}
  \STATE $\tau_j \equiv \{s_t, a_t, s_{t+1}, r_{t+1} \}_j \sim \pi_{\theta}$
  \STATE \codecomment{\# Add data to the buffer}
  \STATE $R \leftarrow R \cup \tau_j$
  \ENDFOR
  \STATE \codecomment{\# Update policy and value functions}
  \FOR {$j$=1,2,..$N_\text{epochs}$}
  \STATE Sample a minibatch of data $r$ from $R$\\
  \STATE $\pi_{\theta}$ $\leftarrow$ PPOUpdate($\pi_{\theta}$, $r$)\\
  \STATE $V_{\phi}$ $\leftarrow$ PPOUpdate($V_{\phi}$, $r$)\\
  \ENDFOR
  \STATE \codecomment{\# Sample all robots uniformly at random initially}
  \IF{$i >= N_{\text{warmup}}$}
  \STATE $P$ $\leftarrow$ samplingProbUpdate(R), according to \Eqref{eq:rb}\\
  \ENDIF
  \ENDFOR
\end{algorithmic}
\end{algorithm}

\begin{table}[!h]
\begin{center}
\begin{tabular}{llr}
\toprule
& \textbf{Hyperparameter} & \textbf{Value}  \\
\midrule
\multirow{21}{*}{PPO} & Discount $\gamma$ & $.99$ \\
&   GAE parameter $\lambda$ & $0.95$ \\
&   PPO clipping parameter $\epsilon$ & $0.2$ \\
&   Policy epochs & $8$ \\
&   Batch size & $5120$ \\
&   Entropy coefficient & $0.01$ \\
&   Reward normalization & Yes \\
&   Reward clipping & $[-10, 10]$ \\
&   Observation normalization & Yes \\
&   Observation clipping & $[-10, 10]$ \\
&   Timesteps per rollout & $2560$\\
&   \# Workers & $16$\\
&   \# Environments & $32$\\
&   Total timesteps & $1 \times 10^8$\\
&   Optimizer    & Adam  \\ 
&   Initial learning rate & $0.0003$ \\
&   Learning rate schedule &  Linear warmup and cosine decay \\
&   Warmup Iterations & $5$ \\
&   Gradient clipping ($l_2$ norm) & $0.5$ \\
&   Clipped value function & Yes \\
&   Value loss coefficient & $0.5$ \\
\midrule

\multirow{6}{*}{Transformer} & Number of layers & $5$ \\
&   Number of attention heads & $1$ \\
&   Embedding dimension & $128$  \\ 
&   Feedforward dimension & $1024$ \\
&   Non linearity function & ReLU \\
&   Dropout & $0.1$ \\

\bottomrule
\end{tabular}
\caption{\textbf{Training hyperparameters.}}
\label{table:ppo_hyper}
\end{center}
\end{table}

\begin{table}[!h]
\begin{center}
\begin{tabular}{lr}
\toprule
\multicolumn{2}{c}{\textbf{Kinematics}}\\
\midrule
\textbf{Variation type}  & \textbf{Value}\\ \hline
Limb radius & $[0.03, 0.05]$ \\
Limb height & $[0.15, 0.45]$\\
\multirow{4}{*}{Joint angles} & $[(-30, 0), (0, 30), (-30, 30),$ \\
& $(-45, 45), (-45, 0), (0, 45),$ \\
& $(-60, 0), (0, 60), (-60, 60)$ \\
& $(-90, 0), (0, 90), (-60, 30) (-30, 60)]$ \\ \\
\midrule
\multicolumn{2}{c}{\textbf{Dynamics}}\\
\midrule
Armature & $[0.1, 2]$\\
Density & $[0.8, 1.2] \times$ limb density\\
Damping & $[0.01, 5.0]$\\
Gear & $[0.8, 1.2] \times$ motor gear\\
\bottomrule
\end{tabular}
\caption{\textbf{Dynamics and kinematics variation parameters.}}
\label{table:dyn_kin_var_parm}
\end{center}
\end{table}

\end{document}